\documentclass[10pt, conference, compsocconf]{IEEEtran}
\ifCLASSINFOpdf
  \usepackage[pdftex]{graphicx}
\else
\fi
\begin{document}
%

\title{DesignGPT: Multi-Agent Collaboration in Design}


\author{
  \IEEEauthorblockN{
    Shiying Ding\textsuperscript{1},
    Xinyi Chen\textsuperscript{2},
    Yan Fang\textsuperscript{3},
    Wenrui Liu\textsuperscript{4},
    Yiwu Qiu\textsuperscript{5},
    Chunlei Chai\textsuperscript{6*}
  }
  \IEEEauthorblockA{
    \textsuperscript{1,2,3,4,6}State Key Lab of CAD\&CG, Zhejiang University, Hangzhou, Zhejiang, China\\
    \textsuperscript{5}zaowuyun.com, Hangzhou, Zhejiang, China\\    
    \textsuperscript{*}Corresponding author\\
    Email: \{chatdsy,cadrenaline,22251360,liuwenrui\}@zju.edu.cn, \\qiuyiwu@zaowuyun.com, dishengchai@126.com
  }
}


%


\maketitle

\begin{abstract}
Generative AI faces many challenges when entering the product design workflow, such as interface usability and interaction patterns. Therefore, based on design thinking and design process, we developed the DesignGPT multi-agent collaboration framework, which uses artificial intelligence agents to simulate the roles of different positions in the design company and allows human designers to collaborate with them in natural language. Experimental results show that compared with separate AI tools, DesignGPT improves the performance of designers, highlighting the potential of applying multi-agent systems that integrate design domain knowledge to product scheme design.

\end{abstract}


%
\IEEEpeerreviewmaketitle

\section{Introduction}
In recent years, the development of artificial intelligence (AI) has provided new possibilities for various aspects of design, including supporting designers' innovative designs\cite{tang2019review}. Design is a complex and creative process, and designers rely on their design thinking and design abilities to solve complex design problems. Due to individual cognitive abilities and experiences, designers often face challenges in the process of conceptualizing innovative solutions. Using generative AI tools is one of the solutions. Image generation AI tools such as MidJourney\cite{2edu} and Stable Diffusion\cite{1SD} mainly generate rich images by inputting text prompts, which can provide inspiration for designers and improve the efficiency of product design. Text generative AI is developing rapidly, and large language model (LLM) such as ChatGPT\cite{ray2023chatgpt} has the ability to understand and generate text, and have been used in design education\cite{2edu} \cite{fangrole} and the generation of design solutions. 

However, the existing generative AI embedded workflow often faces challenges in interface usability and interaction patterns, and there is a natural gap between design thinking and machine thinking. How to make AI better understand design thinking is one of the challenges for designers to interact with AI. Human-centered intelligent product design needs to bridge the gap between design thinking and machine thinking\cite{6sun}.

Design thinking mainly focuses on people, helping designers observe user behavior, explore solutions, and optimize design concepts through means such as brainstorming, social thinking, visual thinking, and prototype practice. These stages require support from different design abilities, including user analysis, sketch design, and CMF (color, material\&Finishing) design. Existing AI design tools rarely consider the entire design process, and switching between different generative AI models can easily interrupt the creator's design thinking.

Design tools based on artificial intelligence continue to emerge in professional software to assist engineering and industrial designers in completing complex manufacturing and design tasks. These tools assume more agency roles than traditional Computer Aided Design tools and are often described as “co-creators”\cite{wang2020human}.

Agents usually refer to entities that exhibit intelligent behavior and have qualities such as autonomy, reactivity, initiative, and social ability\cite{wooldridge1995intelligent}. Compared with the symbolic logic, response or reinforcement learning techniques commonly used by AI agents in the past, Large Language Model (LLM) has shown strong abilities in knowledge acquisition and natural language interaction\cite{GPT4}, and has become an ideal choice for building intelligent agents. Agent\cite{torantulino2023auto-gpt} systems based on large language models have been proven to be feasible simulations of human behavior \cite{li2023camel}\cite{3park}. By introducing standardized operating procedures (SOPs)\cite{4metagpt}, it can effectively reduce the problems of psychedelic, repetitive, incoherent, and invalid feedback in the output of large language models, and has good application prospects\cite{qian2023chatdev}.

We have developed a system DesignGPT, which is designed by multiple AI agents in collaboration with designers. Specifically, we have introduced industry-standardized operating procedures (SOPs) for design to improve efficiency. Through the requirements table, agent roles, and conference rooms, we visualize the operating logic of AI agents for designer participants.

\begin{figure*}[!t]
\centering

  \includegraphics[width=\textwidth]{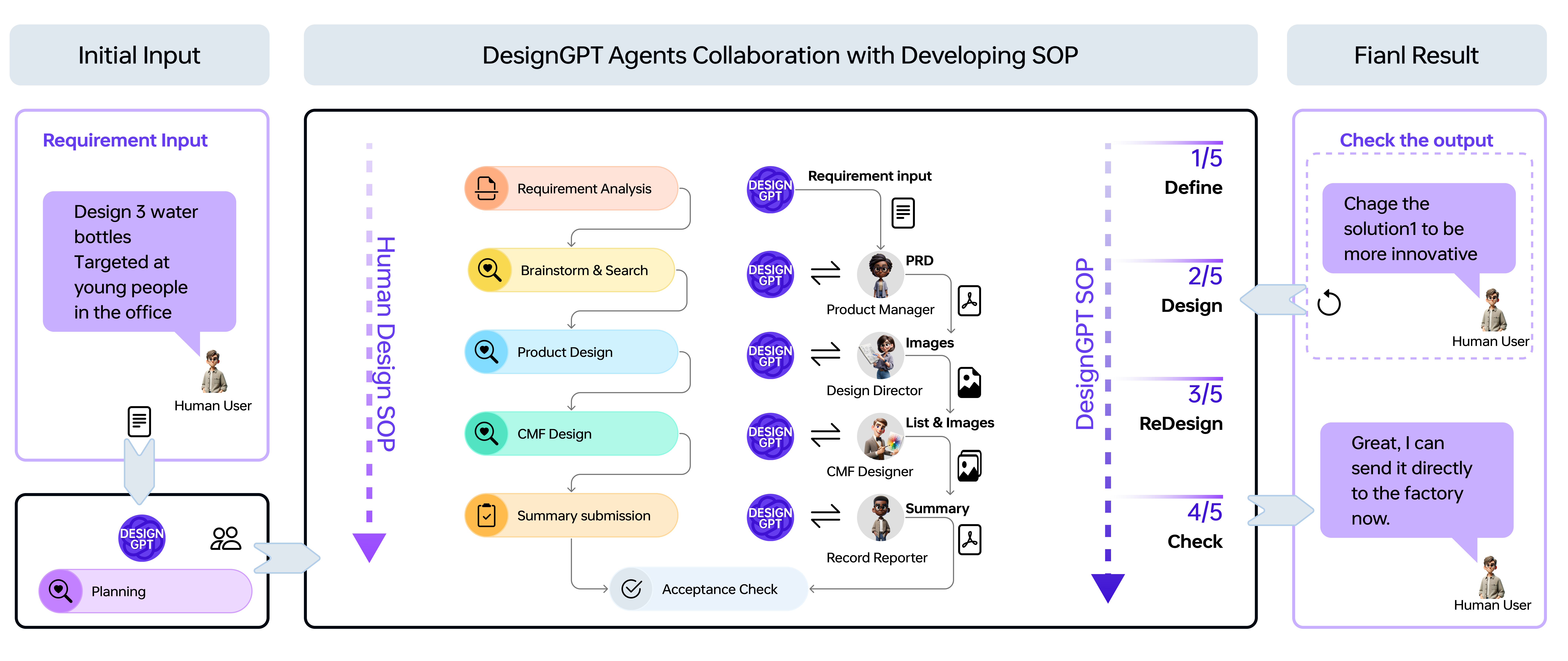}
\caption{System schematic. The system defines the job roles and workflow in the design company. Initially, the designer user needs to input design requirements, such as designing three cups for young people in the office. The DesignGPT system initializes multiple employees in the design company, allowing participants to select roles to start meetings. Employees complete the work in the order of design SOP , and finally output the results. The designer user can check the output of the design process in real time, and intervene by typing midway to make Scheme 1 more innovative.}
\label{fig_1}
\end{figure*}

\section{SYSTEM DESIGN}
We have built the DesignGPT system, a multi-agent collaboration system that incorporates a design thinking knowledge base and the design process, aiming to provide designers with a new method of collaborating with generative AI agents for product design. The system integrates the large language model GPT4\cite{GPT4} and the image generative model Stable Diffusion\cite{1SD}, providing design inspiration through graphic dialogue and supporting designers' innovative design in the conceptual stage. As shown in Fig. \ref{fig_1}.
\subsection{Core components}
The core components of DesignGPT include Requirements Import Form, Role Definition and Selection, and Meeting Room.
\subsubsection{Requirements Import Form}
It mainly presents an interface for filling in design requirements, and users fill in design requirements here. The design requirement will be recorded as an initial prompt, and the design requirement will be set as the meeting topic.

\subsubsection{Role Definition and Selection}
The DesignGPT system initializes multiple employees in the design company, allowing participants to select roles to start meetings. The roles include virtual users, bosses, Product Manager , design directors, CMF designers, scoring recorder, and technical staff. The roles are clearly defined and undertake tasks such as requirement proposal, requirement analysis, design proposal, detailed design, and technical iteration. Virtual users and mainly Product Manager, Design Director, CMF Designers and scoring recorder participate in the final design. In order to facilitate the differentiation of roles among designer participants, we named them ABCDE in order, using random avatars. Product Manager A is responsible for setting the product direction, analyzing the market, and defining product features; Design Director B is responsible for organizing designers and finding design solutions; CMF Designer C is responsible for creating color and style schemes for various scenarios; Recording ReporterD is responsible for Document meetings and assessing proposal feasibility.

\subsubsection{Meeting room}
After filling in the requirements, participants enter the meeting room interface, select employees, and press the Join Meeting Room button to prepare for the meeting. The meeting room chat format is similar to common group chat software. A chat bar is set below the system, and the right side of the chat bar allows uploading pictures to perform picture description and modify color scheme tasks. Role definitions, imported requirements, and historical records will be input to LLM together. Users can supplement the initial design ideas or propose new design requirements according to the selected employee responsibilities. This idea and the initial filled design requirements will be considered in the meeting. After the user enters the content, the agent employees summoned in the meeting room will start the discussion according to their responsibilities, gradually complete the demand analysis, design proposal, detailed design, evaluation report work, and finally output the design plan. During the discussion process, users can output messages through the chat bar to modify the meeting direction.



\subsection{Generative AI settings}
\subsubsection{Image Generation AI settings}
Images are an important part of the design plan. Therefore, the integration of image generation AI needs to ensure the quality of images and the ease of interaction of the system interface. We chose the image generation model based on Stable Diffusion 1.5. Image generation is completed by the role of the design director according to the user's design needs. Sketch is a tool for designers to express design ideas. The process from sketch to final concept map is often time-consuming and laborious, including modifications to product color, material, and process, which we abbreviate as CMF design. The system sets up the role of CMF designer as an agent, and uses Controlnet's canny model to generate design plans. Its input is a picture, and AI automatically extracts the line draft. The line draft is combined with the CMF-related prompt words provided by the agent to generate different CMF design plans.

\subsubsection{Text Generation AI settings}
The text generative model under the system framework can be replaced at will. In order to ensure smooth communication between AI agents and smooth collaboration with designers, we ultimately selected the high-performing model, GPT-4.

The text input to the large language model is called prompt. Our system has designed a module to reconstruct the prompt words input to GPT4. The prompt words generally include role definitions, current skills, output format restrictions, user input, historical records, etc.

The market and user information required for design analysis is updated quickly. Due to the training data for GPT-4 may lag behind the current time, the system has integrated the function of searching for information on the network, and the information obtained from the search will be added as additional text prompts. This function is mainly reflected in the market analysis operation of the Product Manager agent role.

Different roles have different strategies for reconstructing prompt words. For example, in the drawing skills of the design director, the prompt words ultimately used for image generation are comprehensively generated by GPT4 considering the format of Stable Diffusion prompt words, historical records, initial requirements, and current plans.

If only the text model is used, the agent cannot recognize the image, so we use the BLIP\cite{BLIP} model to convert the image uploaded by the user into a text description, and cascade to realize the agent's recognition of the image. Like the network search information function, the text description recognized by BLIP is also added as an additional text prompt to the prompt word submitted to GPT4.

\subsection{Composition of roles}
\begin{figure}[!t]
    \centering
    \includegraphics[width=0.4\textwidth]{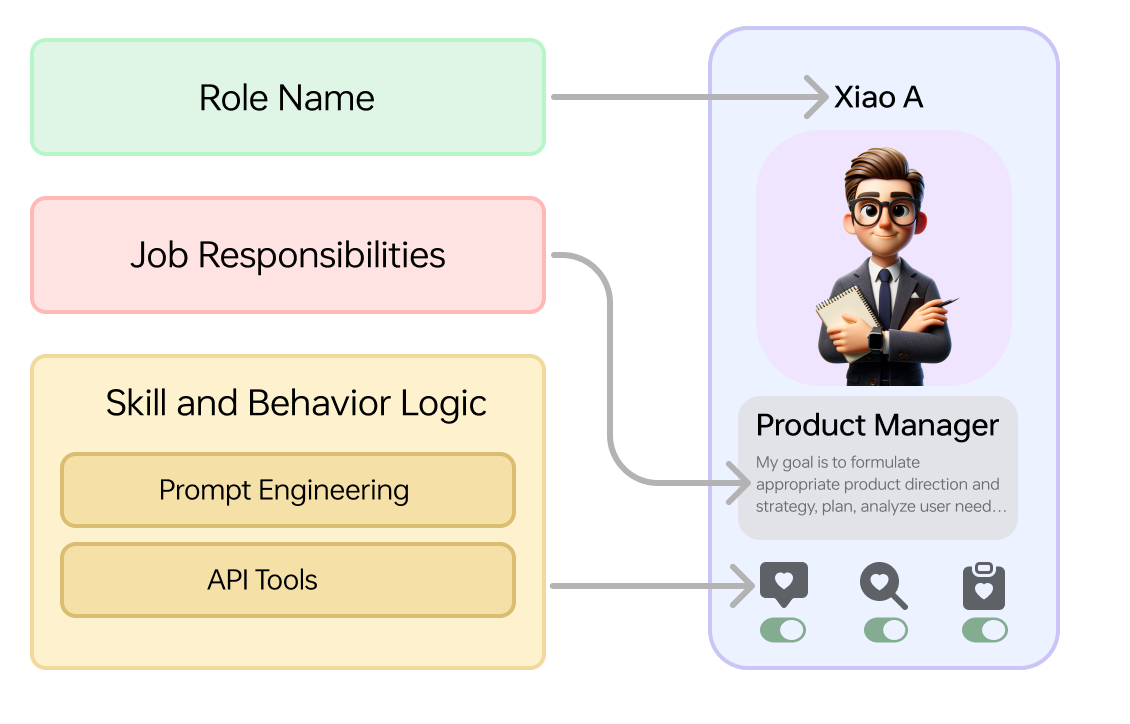}
    \caption{The role card encompasses information such as the role's name (e.g., Xiao A), job responsibilities (e.g., Product Manager), as well as skills and behavior logic, which include prompt engineering and API tools.
}
    \label{fig_role}
\end{figure}
The concept of roles helps designers understand and guides LLM to generate actions that match role characteristics. Therefore, roles under the DesignGPT framework consist of personal information, job responsibilities, and skills. Roles have basic abilities, such as reflection, Knowledge Base, meeting presentation, and status management. Different roles have different skills.

The personal information sets the basic information for the role, such as the name “Xiao A”. The job responsibilities determine the professional knowledge and possible behavioral logic of the role. For example, the goal of Product Manager is to develop appropriate product direction and strategy, planning; competing product analysis, product definition, scene, character portrait; responsible for user research, analyzing user requests, and providing user data support for design.

Skills and logic include the ability of the role and the actions to be followed, which are composed of two parts: prompt engineering and api tools. For example, CMF designer Xiao C needs to import the table for CMF analysis according to the demand, combine the CMF scheme, output the "CMF scheme list ", and utilize a CMF API tool to produce CMF images according to the scheme list.

\section{USER STUDY}
In order to investigate whether the DesignGPT multi-agent collaboration system will change the design process of individual designers and affect the overall design results, and to evaluate the degree of assistance provided by the system in the design process, we conducted an online experiment using inter-group design.

\subsection{Participants}
We recruited 30 design learners for the study, each with an average design learning duration of 2.81 years and an age range between 21 and 24, with an average age of 23. All were college students with similar design proficiency, possessing design experience, computer literacy, exposure to generative AI such as ChatGPT for design.

\subsection{Procedure}
Participants were randomly divided into two groups and designed using one strategy:
\begin{itemize}
  \item \textbf{Strategy 1:} allows designers to create using the AI tools provided by the experiment, specifically the web version of Stable Diffusion (SD) and GPT4.
  \item \textbf{Strategy 2:} allows designers to create web pages using the DesignGPT system, which utilizes the same SD and GPT APIs as Strategy 1.
\end{itemize}
The final output is a design scheme of a product concept. Participants are required to design a “human-centered future intelligent product” in 30 minutes, create a complete design plan, mainly including requirement analysis, concept proposal, specific design presentation, and finally present it in a simple document layout. During the experiment, participants evaluate the product design process themselves, and the system will fully record and export the process document.

\begin{figure}
  \centering
  \includegraphics[width=0.3\textwidth]{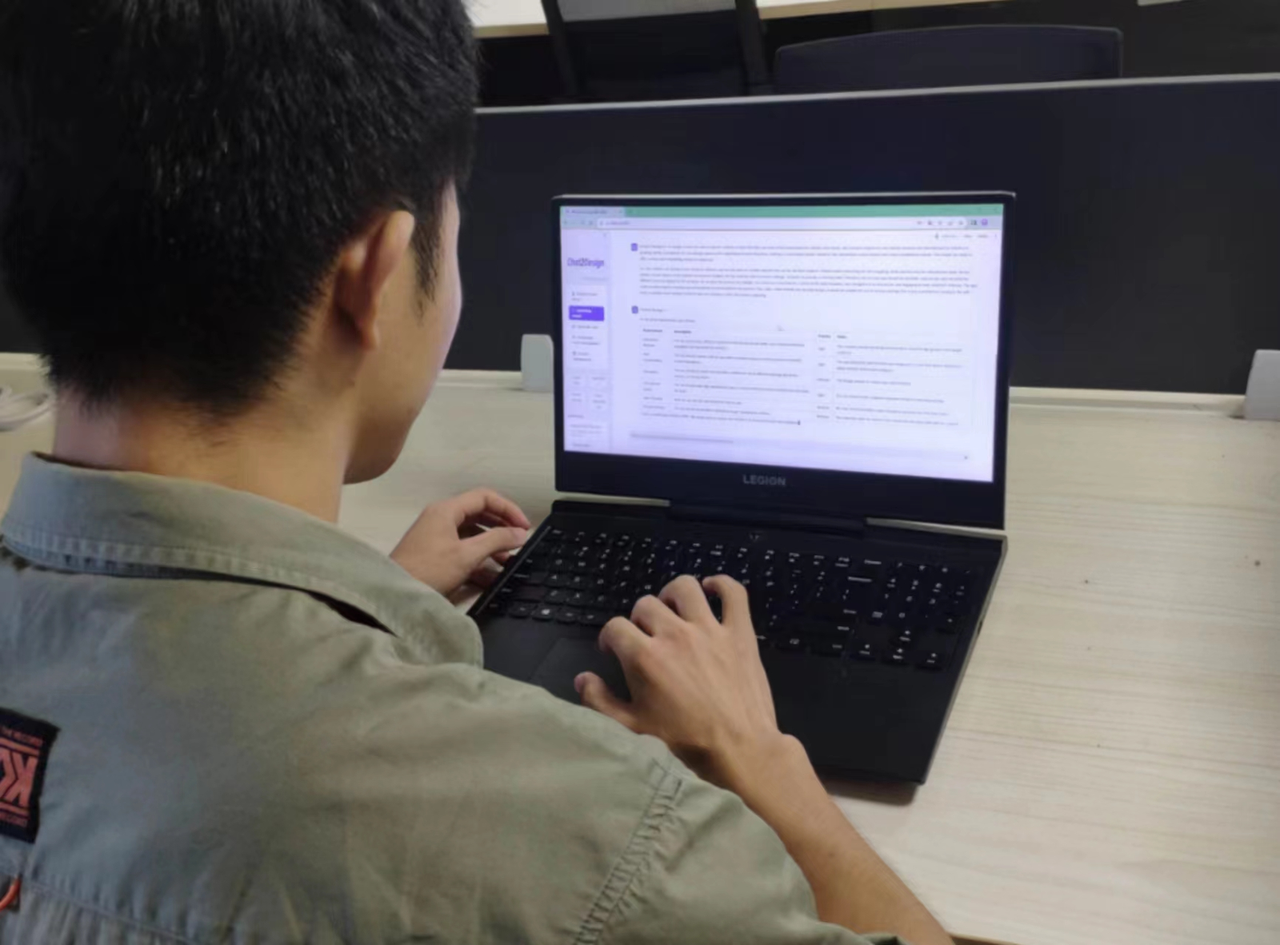}
  \caption{A designer is using DesignGPT system.}
  \label{fig:3}
\end{figure}

\subsection{Evaluation indicators}
In this experiment, we focus on mapping the effectiveness of the design process through the quality of the final plan, so we choose to evaluate the design plan in three dimensions: novelty\cite{chen2019artificial}, completeness, and feasibility\cite{verhaegen2013refinements}. Novelty measures the uniqueness of the design plan, and the evaluation criteria are the novelty compared to existing products or concepts on the market and other concepts generated in this experiment. Feasibility measures the viability of the design plan in technology and the market, as well as the closeness to meeting the requirements of the design summary. completeness measures whether a design idea is fully expressed, and the judgment is based on the richness of details in terms of function, form, material, structure, color, etc.

This evaluation uses a seven-point scale, with 1 point representing the lowest score and 7 points representing the highest score. The evaluators need to score the design scheme in terms of novelty, completeness, and feasibility by viewing the integrated design scheme content and design conclusions.

After the design task, we conducted a 10-15 minute semi-structured interview with each participant about the designer's reaction to the system. The content includes the degree of support and contribution of AI to the design scheme.

\section{Experimental results}
Based on Consensual Assessment Technique (CAT)\cite{amabile1983socialCET}, we invited two experts with 6 years of design experience to independently score each design scheme. The reliability among judges is high, with an inter-group correlation coefficient of 0.90 ($p < 0.001$).

\subsection{quantitative analysis}
\begin{figure}
  \centering
  \includegraphics[width=0.5\textwidth]{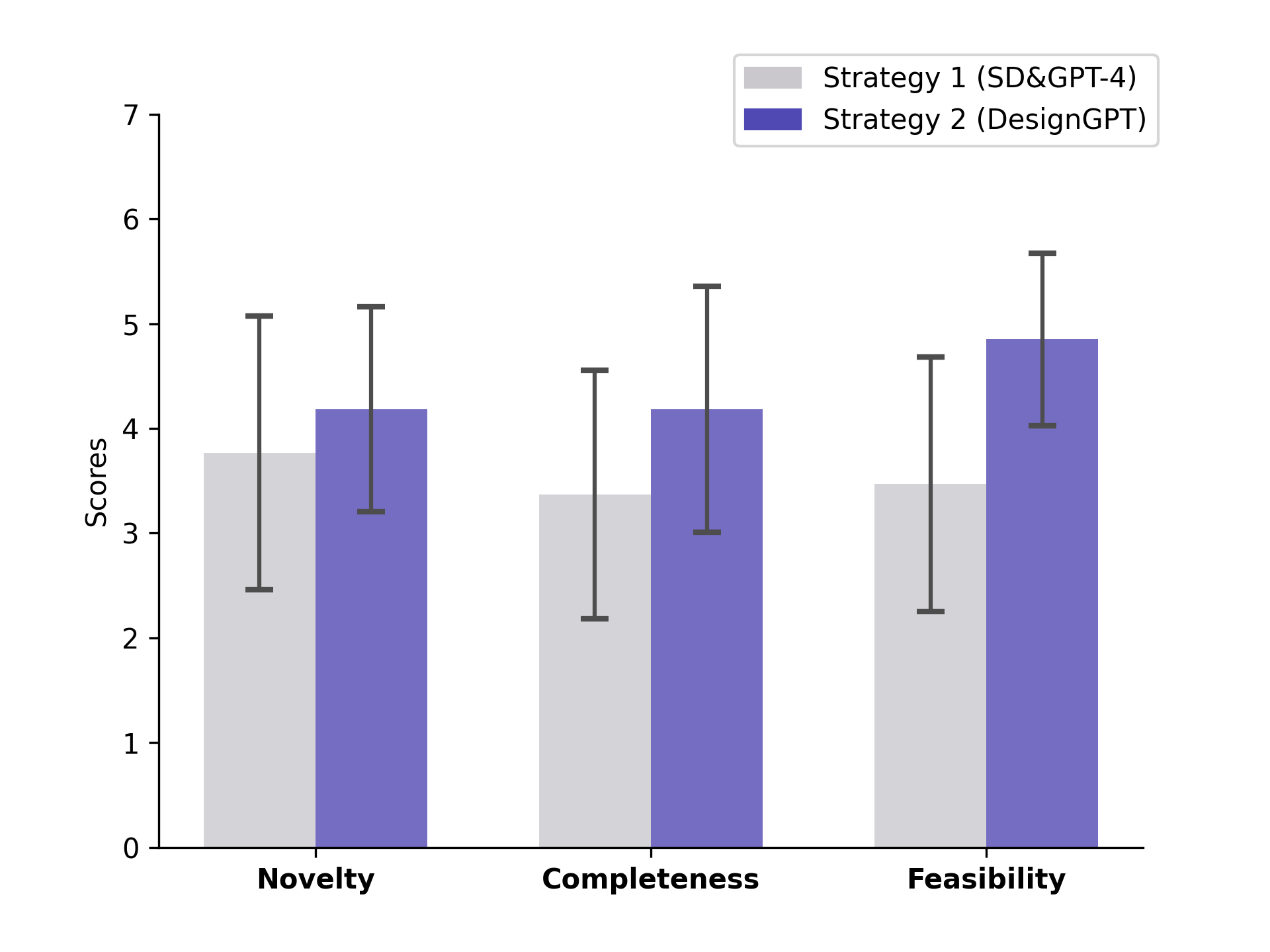}
  \caption{Mean with SD Plot of the designing results. The plot shows the average scores of the participants' design schemes evaluated by experts in the three dimensions of novelty, completeness and feasibility under the two strategies}
  \label{fig:4}
\end{figure}

As shown in Fig. \ref{fig:4}, in terms of three indicators, the mean and median scores of Strategy 2 (DesignGPT) are higher than those of Strategy 1 (SD\&GPT-4). Compared with Strategy 1, Strategy 2 group has an improvement of 11.06\%, 24.26\%, and 39.90\% in novelty, completeness, and feasibility, respectively.Utilizing an independent samples one-tailed t-test (p = 0.05), our analysis reveals that Strategy 2 demonstrates statistically significant advantages over Strategy 1 in terms of completeness and feasibility. Strategy 2 participants' schemes have gained higher overall recognition. The score variance and Standard Deviation of Strategy 2 are slightly lower than those of Strategy 1, indicating that the scheme scores produced by the DesignGPT collaborative system are more concentrated and the evaluators' evaluations are more consistent. 

\subsection{Qualitative Analysis}
Regarding the level of support provided by AI roles in design thinking and program output, Strategy 1 ranges from 5\% to 90\%, and Strategy 2 ranges from 30\% to 80\%. Compared to Strategy 2, Strategy 1 seems to be more likely to have completely AI-led design or designer-led design.
The users gave high praise for the interactive form of the DesignGPT. P5 pointed out that the AI role proficiently uses the design process and design expression, including background analysis, requirement import form, concept screening, scheme expression and scheme scoring, etc. P12 pointed out that DesignGPT deduces the scheme from the perspective of multiple roles, breaking through the problem of having a relatively single perspective in conceiving the scheme in the past.

\section{Conclusions}
Our research developed DesignGPT, a multi-agent-based design collaboration system that supports designers to collaborate with AI agents to complete innovative designs during the conceptual stage. We first conducted an interview study and invited design learners who have tried AI design tools to experiment to understand how designers use AI tools for design and what challenges they encounter during the process. We created  a chat room style interactive website that allows users to input requirements and form collaborative AI agent teams, establishing a new workflow of generative AI with designers. Our results show that the DesignGPT system can effectively support designers' design process. We believe that this research has important theoretical and practical significance for understanding the role of AI in the design process.

\section*{Acknowledgment}
The authors would like to thank Yu Lei and Yuancong Shou for their assistance in implementing experiments and analyzing results. This work was partly supported by “Pioneer” and “Leading Goose” R\&D Programs of Zhejiang[2023C01219].

\newpage


\IEEEtriggeratref{10}



\bibliographystyle{IEEEtran}
\bibliography{IEEEabrv,2}
%






\end{document}